\documentclass[10pt,twocolumn,letterpaper]{article}
\usepackage[pagenumbers]{cvpr}
\usepackage{times}
\usepackage{graphicx}
\usepackage{amsmath,amssymb,booktabs}
\usepackage{textcomp}
\usepackage[breaklinks,colorlinks,allcolors=blue]{hyperref}

\def\confName{CVPR}
\def\confYear{2027}
\title{Coarse Intent, Fine Physics: Returning Latent Forecasts to a Planning Agent}

\author{Kaixuan Yin$^{1*}$\quad Shicheng Yin$^{1*}$\quad Yueyang Ou$^{2}$\quad Hao Li$^{1}$\\
Weixing Chen$^{1}$\quad Ziliang Chen$^{3}$\quad Yang Liu$^{1,4\dagger}$\\
$^{1}$Sun Yat-sen University\quad $^{2}$Tianjin University\quad $^{3}$Peng Cheng Laboratory\quad $^{4}$X-Era AI Lab\\
{\small \{yinkx3, yinshch5, lihao323, chenwx228\}@mail2.sysu.edu.cn,\ \ liuy856@mail.sysu.edu.cn}\\
{\small ouyueyang@tju.edu.cn,\ \ c.ziliang@yahoo.com}\\
{\small Code: \href{https://github.com/HCPLab-SYSU/CIFP}{\textcolor{black}{github.com/HCPLab-SYSU/CIFP}}}}

\makeatletter\@namedef{ver@eso-pic.sty}{2020/10/14}\makeatother
\usepackage{pdfpages}
\begin{document}

\maketitle

{\let\thefootnote\relax\footnotetext{$^{*}$Equal contribution.\quad $^{\dagger}$Corresponding author.}}
\begin{abstract}
A visual agent can decide which side of an object to push and which way it should turn, but it may misjudge where contact will actually carry the object. A learned world model predicts that motion, yet it is usually consulted only through an optimiser's cost, out of the agent's sight. We instead return the model's forecasts to the agent. A correspondence readout turns the frozen world model's predicted features into the motion of points the agent marks on the object, drawn over the current image with a cloud of nearby outcomes; the agent revises its actions and commits exactly the plan it previewed. On whole-trajectory Push-T, a GPT-6 Astra agent solves 10 and 12 of 21 tasks unaided in two deployments. With previews and no optimiser it solves 16 and 21; paired by task, it gains 9 and loses 3 in the first deployment and gains 9 and loses 0 in the second, and all three variants of the tool improve on the bare agent in both. Image-cost MPC seeded with the same agent's proposal solves 3 and 5. On the plans the agent commits, the median step-5 forecast error against simulator ground truth is about 5 pixels with previews alone and 7 with refinement, and wider clouds go with larger errors. The agent supplies coarse intent; the world model, read through correspondence, supplies fine physics.
\end{abstract}

\section{Introduction}

Pushing a T-shaped block into a target pose asks for two things at once. One is intent: which side to contact, which way the block must rotate, which intermediate pose makes the rest easy. Vision-language agents state such intents readily from images. The other is physics: what a particular action sequence will do once contact, friction and rotation take over. A learned world model predicts this~\cite{zhou2024dinowm,yin2026ddpwm}; an agent reasoning from images alone may or may not, and when it does not, it has no way to check.

\begin{figure*}[t]
\centering
\includegraphics[width=\textwidth]{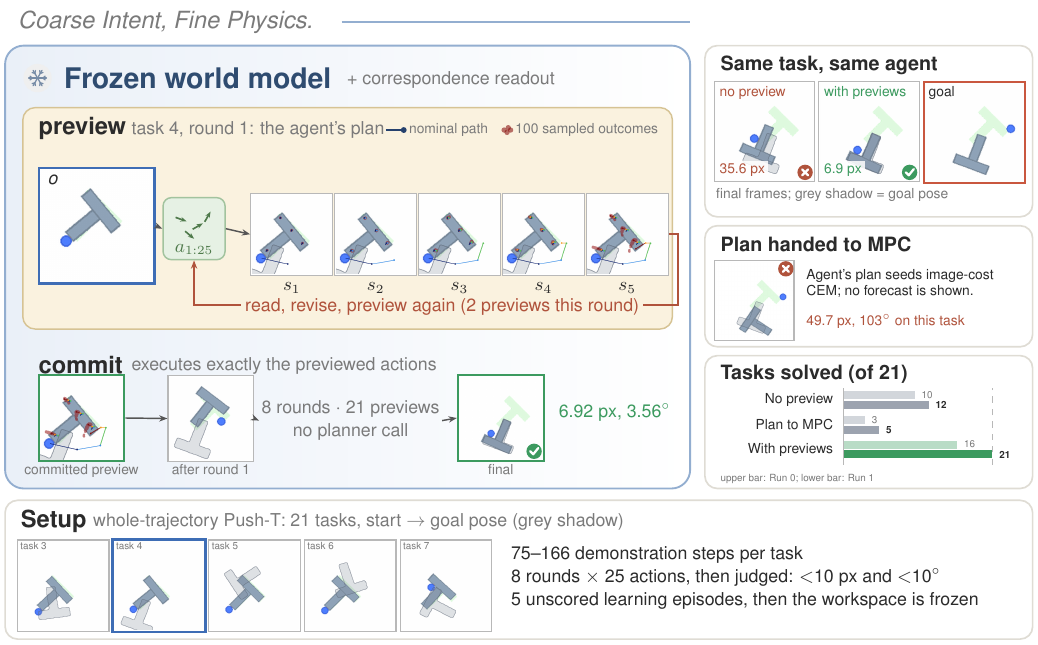}
\caption{Returning a frozen world model's forecasts to the agent, shown on task 4 of Run~1 with real renders. Left: the agent writes a plan; the world model and correspondence readout draw the predicted motion of its marked points over the current image for steps $s_1$--$s_5$, with 100 outcomes (the nominal plan and 99 perturbed variants); the agent reads, revises and previews again, then commits exactly the previewed actions. Right: on the same task, the bare agent ends 35.6\,px from the goal and image-cost MPC seeded with its plan ends 49.7\,px and 103$^\circ$ away, while the agent with previews succeeds (6.9\,px); below, tasks solved out of 21 by these three controllers in both runs. Bottom: the whole-trajectory Push-T setup.}
\label{fig:teaser}
\end{figure*}

The usual way to use a world model is to let an optimiser search actions under a predicted goal distance. The model then contributes physics, but the search has no intent beyond its cost. An agent can propose the starting point for this search, yet it still never sees what the model predicts. We take the opposite route. The agent keeps authority over actions, and the world model becomes a tool that shows it the predicted consequences of what it proposes (Fig.~\ref{fig:teaser}).

The tool needs no future images. The agent marks a few points on the object. For a proposed action sequence, a frozen DDP-WM~\cite{yin2026ddpwm} rolls out predicted DINOv2 features~\cite{oquab2024dinov2}, and a correspondence readout~\cite{jiang2021cotr} locates the same points in each predicted feature map. The agent receives five images, one per predicted step, showing the nominal path of each point and the cloud of outcomes from 99 perturbed variants. It revises and previews again, may ask a local planner to refine towards targets it marks, and finally commits; execution uses exactly the previewed plan.

We test this on whole-trajectory Push-T, where each of the 21 released test trajectories becomes one task spanning the whole demonstration, under a strict 10-pixel and 10-degree criterion. Our main agent, GPT-6 Astra, fails about half of the tasks on its own. Once it can read the model's forecasts, it solves 16 to 21 tasks, in two deployments and with three variants of the tool, and paired per task every variant gains more tasks than it loses in both deployments. Seeding image-cost MPC with the same agent's proposal does much worse.

We contribute (i) an interface that turns a frozen latent world model into an image-space physical preview through query correspondence, with preview-bound commitment; (ii) a complete evaluation of five controllers built on the same predictor, reported per deployment and per task; and (iii) an execution trace of how previews are used.

\section{Forecast Feedback as a Tool}

\subsection{From Latent Features to Image Points}

The predictor is frozen DDP-WM dynamics on DINOv2 patch features. It takes three history frames and predicts five future steps, each consuming a block of five two-dimensional pusher actions. A correspondence readout takes an anchor feature map, a target feature map and a query coordinate in the anchor image, and returns the location of the same object point in the target. It is a COTR-style transformer~\cite{jiang2021cotr} with 1,968,003 parameters. Each frame is resized to $196\times196$ and encoded by DINOv2 ViT-S/14 into 196 patch tokens of dimension 384. A shared linear layer projects the tokens of both frames to 128 dimensions; the 392 tokens, anchor first, receive a learned positional embedding and pass through a four-layer encoder with four heads, feed-forward width 512, GELU, pre-layer normalisation and dropout 0.1, with full self-attention across both frames. A query point is embedded by a fixed sinusoidal encoding of its normalised anchor coordinates, using 32 linear frequencies $k\pi$ with sine and cosine for each axis (128 dimensions), plus a linear embedding of its anchor visibility. A four-layer decoder with the same settings cross-attends from the queries to the encoded tokens; a diagonal mask prevents queries from attending to one another, so each point is read out independently. A two-layer head (128--128--3, GELU) outputs the target coordinates and a visibility score.

The readout is trained on real features only: both inputs are DINOv2 features of recorded frames, and at test time the target input is replaced by the predictor's features, so the readout is applied to predicted feature maps without retraining. The loss is an unweighted L1 on the three outputs. AdamW uses learning rate $10^{-4}$ without a schedule, betas (0.9, 0.999), weight decay 0.01, no augmentation, and batch size 32 on each of eight GPUs. Training was stopped after 49 epochs; the validation-best checkpoint (epoch 47) reaches a normalised L1 of 0.006638, about 1.5 pixels at resolution 224.

\subsection{Points, Previews and Commitment}

The agent first annotates five points in normalised image coordinates: point 0 on the pusher and points 1--4 on the T-shaped object. Annotation returns a labelled image, and a separate confirmation activates the queries. After the first committed action, identities are fixed and the readout tracks the points. The agent may also draw target coordinates, displayed as hollow circles and not given to the world model.

An action file holds five two-dimensional macro commands in $[-1,1]$. Each is the third command of its five-step block; linear interpolation connects successive macro commands, the first two and last two low-level commands are held at the first and last macro values, and the resulting 25 commands are executed at 10\,Hz. A preview evaluates the proposal without executing it and returns five images, one per predicted step, each drawn over the current observation. At step $k$ the image shows the nominal path of every point up to $k$ and the cloud of outcomes at $k$; colours encode time from dark blue to dark red.

Commit executes the most recently previewed snapshot. Editing the file afterwards does not change what is executed; a changed plan must be previewed again. After execution the agent receives the five observed frames with the selected forecast and the readout's matched points overlaid, then plans the next round from the new observation. Simulator pose is used only by the evaluator.

\subsection{Clouds and Optional Refinement}

Let $u_{1:5}$ be the five macro commands and $a=E(u_{1:5})\in\mathbb R^{25\times2}$ the interpolated low-level sequence. With observed features $z_t$ of the current and two previous frames, the frozen predictor rolls out predicted features $\hat z_{t+1},\dots,\hat z_{t+5}$, where step $k$ consumes the low-level commands $a_{5k-4:5k}$. For a marked point with image coordinate $q_j$, the readout $C$ returns its predicted location at step $k$,
\[
\hat p_j^{(k)}(a)=C\big(z_t,\hat z_{t+k},q_j\big),
\]
and the preview draws $\hat p_j^{(1)},\dots,\hat p_j^{(5)}$ as the nominal path. An ordinary cloud preview repeats this for 99 variants $a^{(i)}=a+\epsilon^{(i)}$ with $\epsilon^{(i)}\sim\mathcal N(0,\sigma^2 I)$ and $\sigma=0.01$ native action units, added to every low-level action and dimension and divided by the action standard deviation when expressed in model coordinates. An expanded preview evaluates 300 trajectories at scale 0.02 in macro-action space. Because the predictor is deterministic, the cloud shows the local sensitivity of the learned dynamics to small action changes, not the model's epistemic uncertainty. We summarise its width by the spread
\[
\textstyle S^{(k)}=\frac14\sum_{j=1}^{4}\frac1N\sum_{i=1}^{N}\big\|\hat p_j^{(k)}(a^{(i)})-\bar p_j^{(k)}\big\|,
\]
the mean distance, over the $N=99$ perturbed variants, of the four object points from their mean $\bar p_j^{(k)}$, which Sec.~4 relates to forecast error.

\begin{figure}[t]
\centering
\includegraphics[width=0.78\columnwidth]{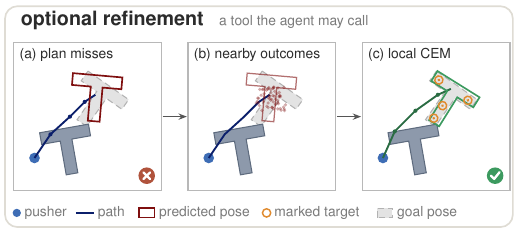}
\caption{Optional refinement (schematic). A proposed path misses the target (a); the expanded preview shows nearby outcomes (b); local CEM towards agent-marked targets returns a plan with its preview (c), which the agent may adopt or discard.}
\label{fig:schematic}
\end{figure}

Where refinement is available, the agent marks targets $g_j$ for the four object points and requests local search initialised from its latest expanded preview. Feasible-first ranking selects 30 elites, whose statistics and best candidate start ten CEM iterations~\cite{rubinstein1999cem} of 300 candidates and 30 elites minimising
\[
J(a)=\sum_{j=1}^{4}\big\|\hat p_j^{(5)}(a)-g_j\big\|+R(a),
\]
where $\hat p_j^{(5)}(a)$ is the predicted endpoint of point $j$ from the current observation and $R$ penalises changes of pairwise point distances beyond tolerance 0.04 with weight 1. Ranking adds $\mathcal N(0,1)$ noise. The refined plan returns with its preview, and the agent decides whether to adopt, edit or discard it (Fig.~\ref{fig:schematic}).

A nominal-feedback variant uses the same images and refinement backend, but its previews show only the unperturbed prediction, and an identical sequence may be previewed once per round. Each test round allows 20 previews, including expanded ones; a preview returned by refinement is additional. During the five learning episodes the limit is five previews per round.

\subsection{One Round of Control}

Putting the pieces together, a round of the cloud-preview controller proceeds as follows. (1) The agent receives the current observation, the goal image and, after the first round, the five observed frames of the previous round with the committed forecast and the tracked points overlaid. (2) It writes five macro commands to its action file. (3) It requests a preview; the service rolls out the proposal and 99 variants through the frozen predictor, reads the five marked points in every predicted feature map, and returns the five step-indexed images. (4) The agent compares the predicted motion with the goal pose and either edits the file and previews again or decides to commit; up to 20 previews are allowed per round. (5) On commit, the service executes exactly the snapshot captured by the latest preview, 25 low-level commands at 10\,Hz. (6) The next round begins from the new observation. The agent never sees latent tensors, never receives simulator state, and is never told whether a preview will succeed; it sees only image-space consequences of its own proposals. With refinement available, step (4) may also include marking targets and requesting a refined plan, which returns as one more preview to accept, edit or discard.

\section{Experimental Protocol}

\subsection{Whole-Trajectory Push-T}

We use the Push-T environment, actions and released test trajectories of DINO-WM~\cite{zhou2024dinowm}, based on~\cite{chi2023diffusion}. A circular pusher moves a T-shaped block. Every test trajectory defines one task, starting from its last frame and targeting the block pose in its first valid frame; for trajectories 14 and 15 the first frame is physically invalid (the pusher overlaps the block, or the block crosses the wall), so the goal is the fourth frame, the first one that is valid together with the five frames after it. The demonstrations connecting these endpoints contain 75--166 low-level steps. Each round executes 25 low-level commands and then observes again; an episode allows eight rounds, at most 200 commands. Agent controllers succeed when the final block position error is below 10 simulator pixels and the angle error below 10 degrees.

\subsection{Controllers}

All controllers use the same frozen predictor. (1) The \emph{bare agent} sees observations and the goal and writes actions without a predictive tool; after each round it receives the current frame without overlays, whereas the feedback controllers receive the five observed frames with the committed forecast and tracked points drawn on them. (2) \emph{Agent-initialised MPC} uses the agent's proposal as the initial mean of image-cost CEM, which minimises the squared distance between predicted final DINO features and goal features (300 candidates, 30 elites, 30 iterations), and executes the result. (3) \emph{Cloud previews} return cloud images and offer no planner. (4) \emph{Nominal feedback} returns nominal previews and offers refinement. (5) \emph{Cloud feedback} returns cloud previews and offers refinement. Three standalone planners use the predictor without an agent: image-cost MPC with zero initial mean, keypoint CEM scoring the readout's predicted object points against goal points with the same rigid-shape term (300/30/30), and spread-penalised CEM, which adds the mean radial spread of perturbed outcomes to the target distance (75 candidates with four perturbations, eight elites, 30 iterations, weight 1, scale 0.01).

\subsection{Learning, Freezing and Deployments}

The agent is GPT-6 Astra at high reasoning effort, running through the Codex command-line agent. Each agent configuration starts from an empty workspace and plays five unscored learning episodes, indexed 101--105 and disjoint from the test tasks, under the same criterion; it may keep notes and reusable procedures. The workspace is then frozen, and each test episode starts independently from the snapshot. No episode teaches another and no weights are updated. Episodes in which the agent exits without acting count as failures.

We report two complete deployments side by side. Run~0 uses an earlier instruction and adapter arrangement, whose wording left the scope of a workspace-writing tool ambiguous. Run~1 uses Codex 0.169.0, seed 1, one final adapter and clarified instructions, and every configuration relearns from scratch. All final episodes passed an independent CPU replay of outcome, criterion, initial state and action continuity.

\subsection{Statistical Analysis}

Each run gives one outcome per controller and task, so the unit of analysis is the task. We compare each feedback controller with the bare agent of the same run by counting tasks it newly solves and tasks it loses, and test the two counts with an exact two-sided sign test, ignoring tasks on which both agree. Within each run we apply Holm's correction~\cite{holm1979} over the three feedback comparisons. We report the two runs separately rather than pooling them, because they share their 21 tasks and differ in pipeline, so they are not independent samples. Success counts are reported in full for every controller, including those that do worse than the bare agent.

\section{Results}

\subsection{Baselines}

The bare agent solves 10 tasks in Run~0 and 12 in Run~1 (Table~\ref{tab:main}). It reads the scene and the goal, yet fails on about half of the tasks, which require several contacts and rotations.

Agent-initialised MPC starts image-cost CEM from the agent's proposal and solves 3 and 5 tasks, seven fewer than the bare agent in each run; in Run~1 it succeeds on tasks 7, 10, 13, 19 and 21. This baseline tells us that this particular optimiser does not turn the agent's proposal into a good controller. It is not a test of optimisation in general: its image-feature cost differs from the point geometry the agent sees, and an agent-seeded keypoint-cost MPC was not run.

For context only, Table~\ref{tab:planners} lists three standalone planners run earlier under their own looser 20\,px/20$^\circ$ criterion, which is not comparable with the agent tables: image-cost MPC with zero initial mean, keypoint CEM and spread-penalised CEM.

\begin{table}[t]\centering\small
\caption{Standalone planners, successes out of 21 in three runs under the looser 20\,px/20$^\circ$ criterion.}
\label{tab:planners}
\begin{tabular}{lcc}\toprule
Planner & Runs 0 / 1 / 2 & Total \\
\midrule
Image-cost MPC & 2 / 2 / 0 & 4/63 \\
Keypoint CEM & 8 / 9 / 7 & 24/63 \\
Spread-penalised CEM & 0 / 0 / 0 & 0/63 \\
\bottomrule\end{tabular}\end{table}

\subsection{Together: The Agent Reads the Forecasts}

\begin{table}[t]\centering\small
\caption{Push-T successes out of 21 under the strict 10\,px/10$^\circ$ criterion in two deployments of the same agent. Each cell is a complete group.}
\label{tab:main}
\resizebox{\columnwidth}{!}{\begin{tabular}{lcc}\toprule
Controller & Run 0 & Run 1 \\
\midrule
Bare agent & 10 (47.6\%) & 12 (57.1\%) \\
Agent-initialised MPC & 3 (14.3\%) & 5 (23.8\%) \\
\midrule
Cloud previews & 16 (76.2\%) & \textbf{21 (100\%)} \\
Nominal feedback + refinement & 18 (85.7\%) & 19 (90.5\%) \\
Cloud feedback + refinement & \textbf{19 (90.5\%)} & 17 (81.0\%) \\
\bottomrule\end{tabular}}\end{table}

When the forecasts are returned to the agent, every variant of the tool solves more tasks than the bare agent, in both deployments (Table~\ref{tab:main}). Cloud previews add 6 and 9 tasks, nominal feedback with refinement adds 8 and 7, and cloud feedback with refinement adds 9 and 5. The best controller reaches 19 of 21 in Run~0 and 21 of 21 in Run~1.

Because all controllers face the same 21 tasks, each run can be compared task by task (Table~\ref{tab:paired}). In both runs, every feedback variant gains more of the bare agent's failures than it loses of its successes. In Run~1 the bare agent fails on tasks 3, 4, 5, 6, 7, 9, 13, 17 and 19; cloud previews solve every one of them and keep all of its successes, and nominal feedback fails only on tasks 4 and 17. In Run~0 the gains are as large or larger but some losses appear: cloud previews gain 9 and lose 3. We test each run separately with an exact two-sided sign test on the discordant tasks and Holm-correct over the three comparisons; three of the six adjusted values are below 0.05, one in Run~0 and two in Run~1. We do not pool the runs, which share their tasks.

\begin{table}[t]\centering\small
\caption{Per-task changes against the bare agent: tasks newly solved / tasks lost, with Holm-adjusted exact sign-test $p$ over the three comparisons within each run.}
\label{tab:paired}
\resizebox{\columnwidth}{!}{\begin{tabular}{lcccc}\toprule
 & \multicolumn{2}{c}{Run 0} & \multicolumn{2}{c}{Run 1} \\
Controller & Gain / loss & $p_\text{Holm}$ & Gain / loss & $p_\text{Holm}$ \\
\midrule
Cloud previews & 9 / 3 & 0.15 & 9 / 0 & 0.012 \\
Nominal feedback + refinement & 10 / 2 & 0.077 & 7 / 0 & 0.031 \\
Cloud feedback + refinement & 10 / 1 & 0.035 & 7 / 2 & 0.18 \\
\bottomrule\end{tabular}}\end{table}

The strongest controller in Run~1 calls no optimiser at all. The cloud-preview agent sees the predicted consequences of its own proposals and chooses among them; the world model is consulted, not put in charge. Across the two runs the three variants have no consistent order: cloud feedback with refinement leads in Run~0 and cloud previews alone in Run~1, and nominal feedback stays within one or two tasks of the best. What the variants share, and what the bare agent and MPC lack, is that the agent reads the model's forecasts before acting. In Run~0, a four-episode rerun of the episodes affected by the ambiguous wording, with the original snapshots and seeds, raises cloud previews from 16 to 18 and leaves the bare agent at 10; Table~\ref{tab:main} keeps the original complete groups.

\subsection{How the Agent Uses a Preview}

\begin{figure*}[t]
\centering
\includegraphics[width=0.9\textwidth]{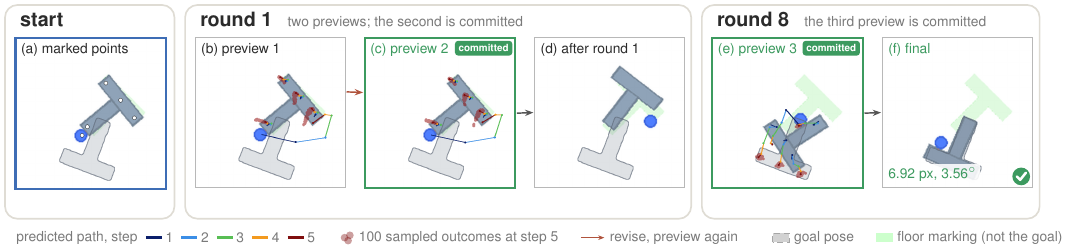}
\caption{Task 4, Run~1, cloud previews: the images returned to the agent. (a) Start with the five marked points. (b) First preview of round one; the cloud shows the 100 step-5 outcomes of the nominal plan and its 99 perturbed variants. (c) The revised plan, previewed and committed. (d) State after round one. (e) The committed third preview of round eight. (f) Final state, 6.92\,px and 3.56$^\circ$ from the goal pose (grey shadow). The pale green floor marking is part of the environment, not the goal.}
\label{fig:case}
\end{figure*}

Task 4 is one of the bare agent's failures. In Run~1 the cloud-preview agent solves it in eight rounds with 2, 2, 2, 3, 2, 4, 3 and 3 previews: 21 previews, no planner calls and 200 low-level commands. In round one it proposes $(0.40,0.08)$, $(0.36,-0.04)$, $(0.04,-0.34)$, $(-0.12,0.04)$, $(-0.14,0.12)$; after the preview it keeps the first three commands, changes the last two to $(-0.115,0.045)$ and $(-0.13,0.115)$, previews again and commits. In round eight it previews three candidates and commits the third, $(-0.14,-0.16)$, $(-0.12,0.13)$, $(-0.075,0.16)$, $(-0.02,0.11)$, $(0.03,0.015)$. Replay gives a final error of 6.92 pixels and 3.56 degrees. The notes the agent saved during learning warn that a tight cloud can still be biased and that rotation tends to be underestimated, so it treats the preview as evidence rather than as a guarantee. Fig.~\ref{fig:case} shows these rounds as the agent saw them.

\subsection{Diagnostics from the Run Records}

The run records store every preview the agent saw, every committed plan, the readout's tracked points at the start of the next round, and the simulator state after every executed round. We use them, without new episodes, to answer three questions that such a system raises.

\textbf{Are the forecasts accurate on the agent's own choices?} We measure forecasts against simulator ground truth: the four object points the agent confirms in the first round are fixed in the block's own frame using the true initial pose, and after each executed round the true block pose, stored in the records and reproduced by CPU replay, maps them back into the 224-pixel image. The error is the mean distance of the four predicted step-5 points from these true points. Over 105 and 100 committed rounds with cloud previews in Run~0 and Run~1, the median error is 5.20 and 5.36 pixels (90th percentile 12.1 and 12.9; 95\% intervals 4.3--5.7 and 5.0--6.3 from resampling whole tasks, since rounds within an episode are dependent), while the predicted displacement of the same points averages 19.5 and 18.1 pixels; with refinement (83 and 89 rounds) the median is 6.70 and 6.73. Taken round by round, the median ratio of error to predicted displacement is 0.32 and 0.34 for cloud previews and 0.45 and 0.34 with refinement. The readout's own tracking error on observed frames is 2.6--3.3\,px (Table~\ref{tab:diag}). These counts include each committed round that is followed by another round; adding the final round of every episode gives 124, 121, 104 and 110 rounds, consistent with the mean rounds in Table~\ref{tab:cost}, and medians of 5.2, 5.2, 6.0 and 6.5\,px. Episodes in which the agent committed nothing contribute no rounds.

\begin{figure*}[t]
\centering
\includegraphics[width=\textwidth]{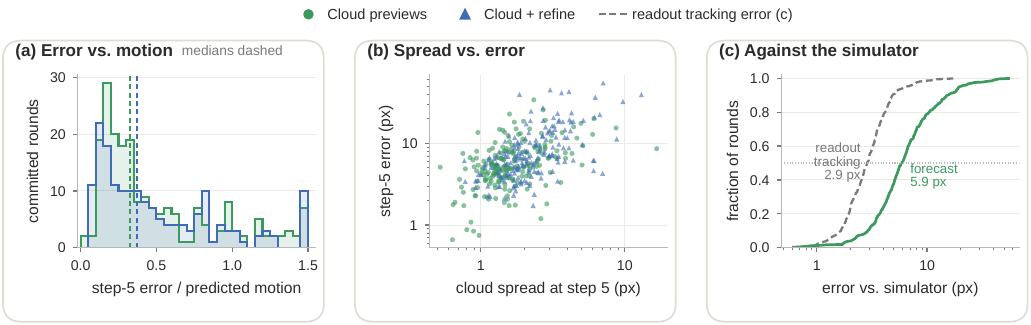}
\caption{Forecasts on the plans the agent committed, against simulator ground truth, pooled over both runs (205 rounds with cloud previews, 172 with cloud + refinement). (a) Step-5 forecast error divided by the predicted displacement of the same object points; values above 1.5 are collected in the last bin and mostly come from rounds with very small predicted motion. (b) Cloud spread at step 5 against the step-5 error, both in 224-pixel image units; wider clouds go with larger errors. (c) Cumulative distributions of the forecast error and of the readout's own tracking error on observed frames (medians 5.88 and 2.86\,px over 377 rounds), so the forecast error is not an artefact of the readout.}
\label{fig:diag}
\end{figure*}

\textbf{Does the cloud mean anything?} Across committed previews, the spread of the cloud at step 5 is positively rank-correlated with the ground-truth error of the nominal forecast: Spearman 0.56 and 0.36 for cloud previews and 0.61 and 0.58 with refinement. This is not merely a proxy for large motions: the predicted displacement itself correlates less with the error (0.31, 0.26, 0.13, 0.22), and with displacement partialled out the spread keeps a rank correlation of 0.52, 0.32, 0.60 and 0.55. Task-resampled 95\% intervals for these correlations stay above zero (lowest 0.14, Run~1 cloud previews). A locally sensitive forecast is thus a less reliable one; the agent's notes say it uses this, but our records do not show that it does so effectively. Fig.~\ref{fig:diag} shows both distributions over all committed rounds of the two runs, and Table~\ref{tab:diag} lists the statistics per run and controller.

\begin{table}[t]\centering\small
\caption{Forecasts on committed rounds against simulator ground truth, per run and controller: number of rounds, median and 90th-percentile step-5 error (224-pixel units), median ratio of error to predicted displacement, Spearman correlation between cloud spread and error, and the readout's own median tracking error on the observed frames.}
\label{tab:diag}
\resizebox{\columnwidth}{!}{\begin{tabular}{llcccccc}\toprule
Run & Controller & Rounds & Median & P90 & Ratio & $\rho_\text{spread}$ & Readout \\
\midrule
0 & Cloud previews & 105 & 5.20 & 12.1 & 0.32 & 0.56 & 2.68 \\
0 & Cloud + refine & 83 & 6.70 & 20.2 & 0.45 & 0.61 & 2.64 \\
1 & Cloud previews & 100 & 5.36 & 12.9 & 0.34 & 0.36 & 2.75 \\
1 & Cloud + refine & 89 & 6.73 & 19.1 & 0.34 & 0.58 & 3.33 \\
\bottomrule\end{tabular}}\end{table}

\begin{figure*}[t]
\centering
\includegraphics[width=\textwidth]{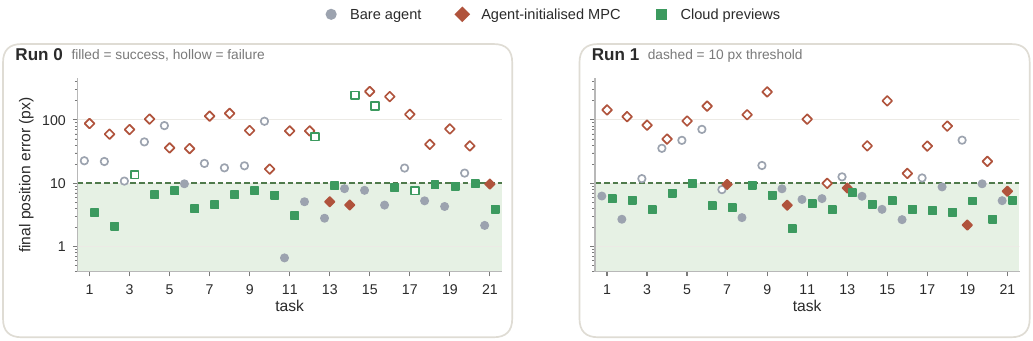}
\caption{Final block position error on each of the 21 tasks (log scale, simulator pixels), from the independent replay, for the bare agent, agent-initialised MPC and cloud previews in both runs. The dashed line is the 10\,px position threshold. Filled markers are successes and hollow markers failures (success also needs an angle error below 10$^\circ$). Cloud previews bring both near and distant failures of the bare agent below the line, while MPC ends far away on most tasks.}
\label{fig:finalerr}
\end{figure*}

\textbf{How far do failures miss, and at what agent cost?} The bare agent's failures are often close: 5 of its 11 failures in Run~0 and 5 of 9 in Run~1 end within 20\,px and 20$^\circ$, with medians of 20.5\,px/7.7$^\circ$ and 19.0\,px/11.0$^\circ$. Agent-initialised MPC fails far away, with median errors of 69\,px/73$^\circ$ and 89\,px/90$^\circ$. Previews repair both kinds of failure. In Run~1, cloud previews solve all five of the bare agent's near misses and all four of its distant failures (35.6--70.8\,px). In Run~0 they solve all six distant failures and three of the five near misses. The three tasks cloud previews lose in Run~0 (12, 14 and 15) end far from the goal (54--243\,px), so the losses are not narrow misses either. The tested MPC, by contrast, discards the agent's coarse progress: its failures end far away even on tasks the bare agent nearly solved. Fig.~\ref{fig:finalerr} plots the final position error of these three controllers on every task, and Table~\ref{fig:pertask} lists every outcome of all five.

\begin{table*}[t]\centering\scriptsize
\setlength{\tabcolsep}{3.2pt}
\caption{Every outcome: success ($\bullet$) or failure ($\circ$) on each of the 21 tasks, for all five controllers in both runs, from the independent replay. For Run~0 cloud + refinement task 12 the run summary lacked the fourth-round actions; replaying all rounds with the actions stored by that run passes every check and ends at 7.94\,px and 8.86$^\circ$, without rerunning the episode.}
\label{fig:pertask}
\begin{tabular}{ll*{21}{c}c}\toprule
 & Controller & 1 & 2 & 3 & 4 & 5 & 6 & 7 & 8 & 9 & 10 & 11 & 12 & 13 & 14 & 15 & 16 & 17 & 18 & 19 & 20 & 21 & Total \\
\midrule
Run 0 & Bare agent & $\circ$ & $\circ$ & $\circ$ & $\circ$ & $\circ$ & $\bullet$ & $\circ$ & $\circ$ & $\circ$ & $\circ$ & $\bullet$ & $\bullet$ & $\bullet$ & $\bullet$ & $\bullet$ & $\bullet$ & $\circ$ & $\bullet$ & $\bullet$ & $\circ$ & $\bullet$ & 10 \\
Run 0 & Agent-init.\ MPC & $\circ$ & $\circ$ & $\circ$ & $\circ$ & $\circ$ & $\circ$ & $\circ$ & $\circ$ & $\circ$ & $\circ$ & $\circ$ & $\circ$ & $\bullet$ & $\bullet$ & $\circ$ & $\circ$ & $\circ$ & $\circ$ & $\circ$ & $\circ$ & $\bullet$ & 3 \\
Run 0 & Cloud previews & $\bullet$ & $\bullet$ & $\circ$ & $\bullet$ & $\bullet$ & $\bullet$ & $\bullet$ & $\bullet$ & $\bullet$ & $\bullet$ & $\bullet$ & $\circ$ & $\bullet$ & $\circ$ & $\circ$ & $\bullet$ & $\circ$ & $\bullet$ & $\bullet$ & $\bullet$ & $\bullet$ & 16 \\
Run 0 & Nominal + refine & $\bullet$ & $\bullet$ & $\bullet$ & $\bullet$ & $\bullet$ & $\bullet$ & $\bullet$ & $\circ$ & $\bullet$ & $\bullet$ & $\bullet$ & $\bullet$ & $\bullet$ & $\circ$ & $\circ$ & $\bullet$ & $\bullet$ & $\bullet$ & $\bullet$ & $\bullet$ & $\bullet$ & 18 \\
Run 0 & Cloud + refine & $\bullet$ & $\bullet$ & $\bullet$ & $\bullet$ & $\bullet$ & $\bullet$ & $\bullet$ & $\circ$ & $\bullet$ & $\bullet$ & $\bullet$ & $\bullet$ & $\bullet$ & $\bullet$ & $\circ$ & $\bullet$ & $\bullet$ & $\bullet$ & $\bullet$ & $\bullet$ & $\bullet$ & 19 \\
\midrule
Run 1 & Bare agent & $\bullet$ & $\bullet$ & $\circ$ & $\circ$ & $\circ$ & $\circ$ & $\circ$ & $\bullet$ & $\circ$ & $\bullet$ & $\bullet$ & $\bullet$ & $\circ$ & $\bullet$ & $\bullet$ & $\bullet$ & $\circ$ & $\bullet$ & $\circ$ & $\bullet$ & $\bullet$ & 12 \\
Run 1 & Agent-init.\ MPC & $\circ$ & $\circ$ & $\circ$ & $\circ$ & $\circ$ & $\circ$ & $\bullet$ & $\circ$ & $\circ$ & $\bullet$ & $\circ$ & $\circ$ & $\bullet$ & $\circ$ & $\circ$ & $\circ$ & $\circ$ & $\circ$ & $\bullet$ & $\circ$ & $\bullet$ & 5 \\
Run 1 & Cloud previews & $\bullet$ & $\bullet$ & $\bullet$ & $\bullet$ & $\bullet$ & $\bullet$ & $\bullet$ & $\bullet$ & $\bullet$ & $\bullet$ & $\bullet$ & $\bullet$ & $\bullet$ & $\bullet$ & $\bullet$ & $\bullet$ & $\bullet$ & $\bullet$ & $\bullet$ & $\bullet$ & $\bullet$ & 21 \\
Run 1 & Nominal + refine & $\bullet$ & $\bullet$ & $\bullet$ & $\circ$ & $\bullet$ & $\bullet$ & $\bullet$ & $\bullet$ & $\bullet$ & $\bullet$ & $\bullet$ & $\bullet$ & $\bullet$ & $\bullet$ & $\bullet$ & $\bullet$ & $\circ$ & $\bullet$ & $\bullet$ & $\bullet$ & $\bullet$ & 19 \\
Run 1 & Cloud + refine & $\bullet$ & $\bullet$ & $\bullet$ & $\bullet$ & $\bullet$ & $\bullet$ & $\bullet$ & $\bullet$ & $\circ$ & $\bullet$ & $\bullet$ & $\bullet$ & $\circ$ & $\circ$ & $\circ$ & $\bullet$ & $\bullet$ & $\bullet$ & $\bullet$ & $\bullet$ & $\bullet$ & 17 \\
\bottomrule\end{tabular}\end{table*} Table~\ref{tab:cost} reports the agent's cost per episode. Cloud previews raise the bare agent's input tokens by a factor of 3.7 and 4.6 and its time by 3.7 and 1.8, but take less time than the tested MPC in both runs. The feedback controllers also use fewer of the eight available rounds: 5.9 and 5.8 rounds per episode for cloud previews and 4.6 to 5.2 with refinement, against 6.5 and 7.5 for the bare agent and about 7 for MPC. Their extra cost goes into previews within a round rather than into more executed rounds, and most of it is input tokens.

\begin{table*}[t]\centering\small
\caption{Mean agent cost per episode, Run~0 / Run~1: wall-clock minutes, executed rounds (of 8), model requests, tool calls, input and output tokens, previews and refinements requested. All values come from the run records; agent inference is metered per episode.}
\label{tab:cost}
\begin{tabular}{lcccccccc}\toprule
Controller & Minutes & Rounds & Requests & Tool calls & Input tok.\ (M) & Output tok.\ (k) & Previews & Refines \\
\midrule
Bare agent & 5.0 / 7.9 & 6.5 / 7.5 & 20.5 / 16.4 & 35.1 / 37.0 & 0.33 / 0.25 & 5.4 / 5.7 & -- & -- \\
Agent-init.\ MPC & 24.6 / 18.9 & 7.1 / 7.2 & 58.4 / 49.2 & 42.4 / 37.1 & 1.03 / 0.83 & 6.1 / 6.1 & -- & -- \\
Cloud previews & 18.3 / 14.3 & 5.9 / 5.8 & 48.4 / 47.2 & 80.8 / 79.9 & 1.21 / 1.16 & 8.1 / 8.6 & 17.3 / 16.7 & -- \\
Nominal + refine & 14.6 / 13.4 & 5.0 / 4.6 & 37.7 / 38.6 & 74.2 / 79.8 & 0.98 / 1.07 & 7.7 / 8.3 & 18.7 / 19.5 & 1.8 / 2.3 \\
Cloud + refine & 23.2 / 20.5 & 5.0 / 5.2 & 64.2 / 72.7 & 93.9 / 116.0 & 1.83 / 2.37 & 9.2 / 11.2 & 19.7 / 26.5 & 1.8 / 2.7 \\
\bottomrule\end{tabular}\end{table*}

On the world-model side, one trajectory expansion predicts a full five-step horizon: a cloud preview uses 100, a nominal preview 1, an expanded preview 300, a refinement 3,000 plus its returned preview, and one MPC round 9,000. Mean expansions per episode were 1,667--1,729 for cloud previews, 5,828--7,579 for nominal feedback with refinement and 64,260--65,160 for agent-initialised MPC across the two runs.

\section{Discussion}

\textbf{What the agent gains from a preview.} Episodes fail either near the goal or far off course; the bare agent produces both, the agent with previews far less often. A preview tells the agent, before it acts, whether a push will rotate the block the intended way, and the trace in Fig.~\ref{fig:case} shows the agent changing its last commands in response.

\textbf{When previews do not help.} The losses concentrate on tasks 14 and 15, whose goals come from the fourth demonstration frame and which the bare agent solves in both runs (Table~\ref{fig:pertask}). On these two tasks the forecasts committed by the cloud controller with refinement were unusually poor against ground truth, with per-task median errors of 7--21\,px and 0.78--0.89 of the predicted displacement, against 0.34--0.45 over all its rounds; the agent acted on a biased model, and a local optimiser can follow such a model into a confident but wrong plan. This explains only the refinement losses: cloud previews also lose tasks 14 and 15 in Run~0 without any optimiser, and our records do not isolate why.

\textbf{What the cloud adds.} Clouds do not solve more tasks than nominal feedback; as a sensitivity display, their spread predicts forecast error better than the predicted motion does, so they indicate how far to trust the nominal path.
\textbf{Why the tested MPC fails.} Agent-initialised MPC keeps the agent's proposal only as a starting mean and then optimises a feature distance to the goal image over a five-step horizon. On long tasks that distance may favour moves that look closer in feature space but undo the agent's progress.

\section{Related Work}

\textbf{World models for control.} Predicting visual consequences for planning has a long history~\cite{finn2017foresight,ebert2018foresight}, including latent dynamics for planning and reinforcement learning~\cite{hafner2019planet,hafner2020dreamer,hansen2022tdmpc,hansen2024tdmpc2}. Joint-embedding models predict learned representations rather than pixels~\cite{lecun2022path,assran2025vjepa2}; DINO-WM plans on frozen DINOv2 features~\cite{zhou2024dinowm} and DDP-WM separates sparse interaction dynamics from the static background~\cite{yin2026ddpwm}. They are usually consumed through sampling-based optimisers~\cite{rubinstein1999cem,pinneri2020icem,williams2017mppi}, with ensembles or penalties for uncertainty~\cite{chua2018pets,webster2021risk}. Keypoints serve as states for predictive control~\cite{manuelli2020keypoints}, and correspondence transformers map points across images~\cite{jiang2021cotr}. We keep the predictor fixed and change who reads its output.

\textbf{Agents and robot action.} Language and vision-language agents ground plans in affordances, write policy code, compose value maps or keypoint constraints, and select actions through visual prompts~\cite{ahn2022saycan,liang2023cap,huang2023voxposer,huang2024rekep,nasiriany2024pivot}. Closest to us, VLMPC samples actions, predicts future videos and asks a vision-language model to score them~\cite{zhao2024vlmpc}; FOREWARN aligns a VLM to a world model's latent foresight to choose among actions proposed by a policy~\cite{wu2025forewarn}; KUDA has a VLM specify keypoint targets that learned dynamics and an optimiser then reach~\cite{liu2025kuda}. In each, a sampler, policy or optimiser produces the actions; our agent writes them, sees their predicted consequences as point geometry, revises, and commits what it previewed. Mark-based visual prompting grounds such agents in image points~\cite{yang2023som,liu2024moka}, and tool use, reflection and language feedback let them revise after acting~\cite{huang2022innermonologue,yao2023react,shinn2023reflexion}; our tool returns predicted physics before execution.

\section{Limitations and Future Work}

\textbf{Scope.} The evidence comes from one simulated task family with 21 tasks and one agent model, GPT-6 Astra. The two runs reuse these tasks and differ in their pipeline, so they show a consistent direction rather than an independent replication.

\textbf{Isolating the forecast.} Feedback gives the agent more than physics: it also adds interaction, images, richer post-round frames and chances to reconsider, and each configuration learns its own procedures. The control that would separate these is an agent with the same preview loop whose previews carry no information about the proposal, for instance shuffled forecasts from other proposals or a constant-velocity extrapolation of the marked points. A nominal-preview controller without refinement would likewise isolate the cloud display. We did not run these, so our claim is about the complete preview tool, not about the share due to physical prediction alone.

\textbf{Stronger optimiser baselines.} An agent-seeded keypoint-cost MPC towards agent-marked targets, under the same strict criterion and round budget, would test whether the gain requires the agent to read the forecast or only a better objective.

\textbf{Forecast quality.} Ground-truth errors are measured only on committed plans, so they describe the forecasts the agent chose to trust, not forecasts on arbitrary actions; the spread measures sensitivity, not epistemic uncertainty, and is not calibrated. There is no real-robot result and no compute-matched comparison.

\section{Conclusion}

Reading a frozen world model's forecasts, a GPT-6 Astra agent solved 16 to 21 of 21 Push-T tasks instead of about half. The agent supplies coarse intent; the consulted world model, fine physics.

\label{body:end}

\clearpage
\includepdf[pages=-]{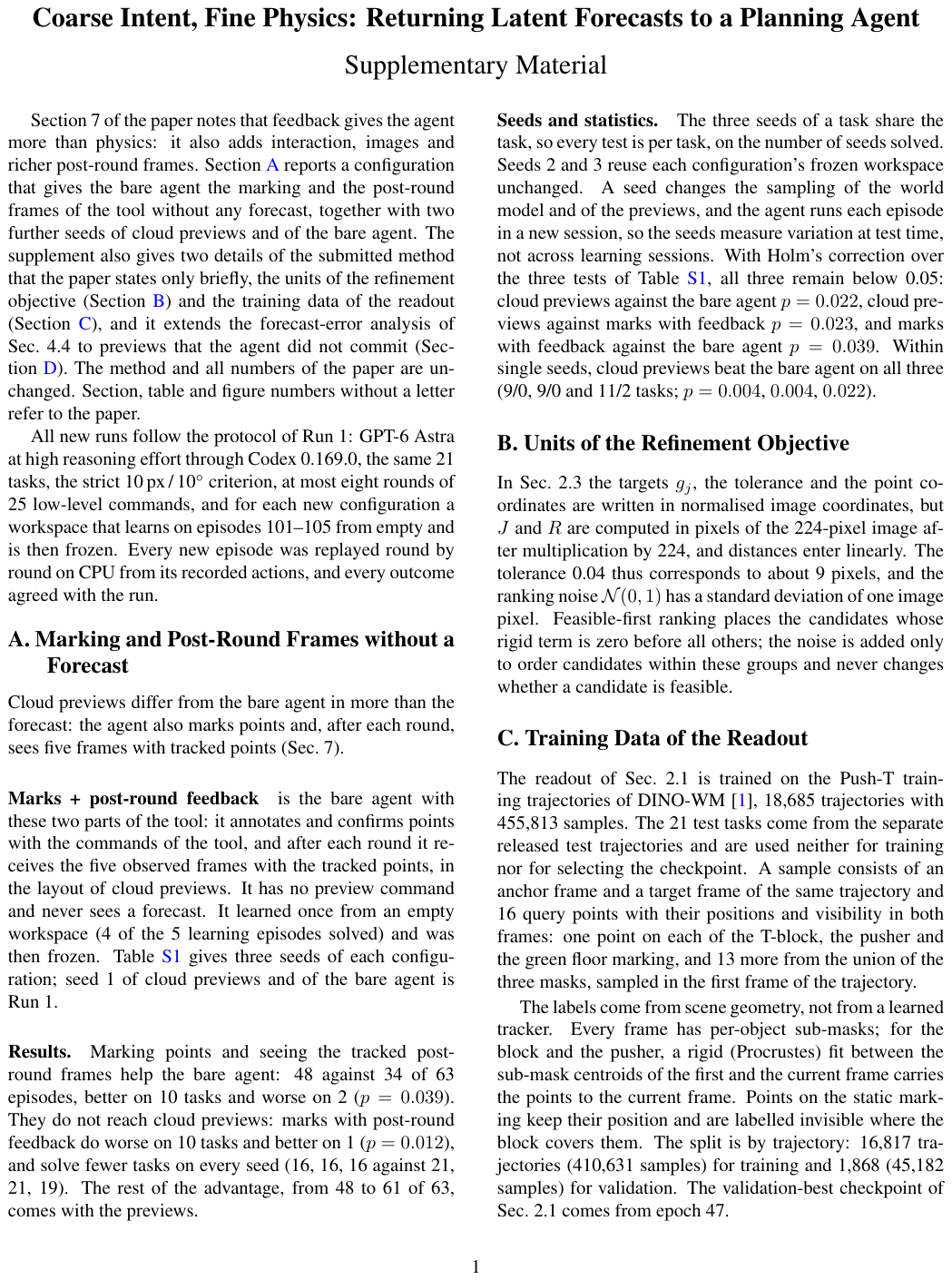}
\end{document}